\documentclass[letterpaper]{article} 
\usepackage{aaai24}  
\usepackage{times}  
\usepackage{helvet}  
\usepackage{courier}  
\usepackage[hyphens]{url}  
\usepackage{graphicx} 
\usepackage{natbib}  
\usepackage{caption} 
\usepackage{algorithm}
\usepackage{algorithmic}
\usepackage{bm}
\usepackage{svg}
\usepackage{multirow}
\usepackage{booktabs}
\usepackage{colortbl}
\usepackage{diagbox}
\usepackage{amssymb}
\usepackage{caption}
\usepackage{subcaption}
\usepackage{pifont}
\usepackage{caption}
\usepackage{fdsymbol}
\usepackage{color}
\newcommand{\cmark}{\ding{51}}%

\definecolor{lightblut}{RGB}{0,162,232}

\newlength\savewidth\newcommand\shline{\noalign{\global\savewidth\arrayrulewidth\global\arrayrulewidth 1pt}\hline\noalign{\global\arrayrulewidth\savewidth}}

\usepackage{newfloat}
\usepackage{listings}
\DeclareCaptionStyle{ruled}{labelfont=normalfont,labelsep=colon,strut=off} 
\floatstyle{ruled}
\newfloat{listing}{tb}{lst}{}
\floatname{listing}{Listing}
\title{DPA-P2PNet: Deformable Proposal-aware P2PNet for Accurate Point-based Cell Detection}
\author {
	Zhongyi Shui\textsuperscript{\rm 1,2}\equalcontrib,
	Sunyi Zheng\textsuperscript{\rm 2}\equalcontrib,
	Chenglu Zhu\textsuperscript{\rm 2},
	Shichuan Zhang\textsuperscript{\rm 1,2},
	Xiaoxuan Yu\textsuperscript{\rm 1,2},
	Honglin Li\textsuperscript{\rm 1,2},
	Jingxiong Li\textsuperscript{\rm 1,2},
	Pingyi Chen\textsuperscript{\rm 1,2},
	Lin Yang\textsuperscript{\rm 2}\thanks{Corresponding author.}
}

\affiliations {
	\textsuperscript{\rm 1}College of Computer Science and Technology, Zhejiang University\\
	\textsuperscript{\rm 2}School of Engineering, Westlake University\\
	\{shuizhongyi,yanglin\}@westlake.edu.cn
}

\usepackage{bibentry}

\begin{document}

\maketitle

\begin{abstract}
	Point-based cell detection (PCD), which pursues high-performance cell sensing under low-cost data annotation, has garnered increased attention in computational pathology community. Unlike mainstream PCD methods that rely on intermediate density map representations, the Point-to-Point network (P2PNet) has recently emerged as an end-to-end solution for PCD, demonstrating impressive cell detection accuracy and efficiency. Nevertheless, P2PNet is limited to decoding from a single-level feature map due to the scale-agnostic property of point proposals, which is insufficient to leverage multi-scale information. Moreover, the spatial distribution of pre-set point proposals is biased from that of cells, leading to inaccurate cell localization. To lift these limitations, we present DPA-P2PNet in this work. The proposed method directly extracts multi-scale features for decoding according to the coordinates of point proposals on hierarchical feature maps. On this basis, we further devise deformable point proposals to mitigate the positional bias between proposals and potential cells to promote cell localization. Inspired by practical pathological diagnosis that usually combines high-level tissue structure and low-level cell morphology for accurate cell classification, we propose a multi-field-of-view (mFoV) variant of DPA-P2PNet to accommodate additional large FoV images with tissue information as model input. Finally, we execute the first self-supervised pre-training on immunohistochemistry histopathology image data and evaluate the suitability of four representative self-supervised methods on the PCD task. Experimental results on three benchmarks and a large-scale and real-world interval dataset demonstrate the superiority of our proposed models over the state-of-the-art counterparts. Codes and pre-trained weights will be available.
\end{abstract}

\section{Introduction}
Identifying various types of cells such as tumor cells, lymphocytes, and fibroblasts in histopathology whole slide images (WSIs) is crucial for numerous downstream tasks including tumor microenvironment analysis \cite{jiao2021deep}, cancer diagnosis \cite{cheng2022artificial} and prognosis \cite{howard2019exploring}. Predominant cell detection methodologies embrace a instance segmentation paradigm. Although these approaches showcase impressive capability in capturing intricate details of cell morphology, their training requires a formidable investment of valuable resources due to the laborious nature of cell mask annotation. As a matter of fact, the expensive annotation has long plagued the advancement and application of mask-based cell detection models.

To reduce the annotation cost while maintaining sufficient clinical support, point-based cell detection (PCD) has emerged as a promising and rapidly evolving frontier in computational pathology \cite{zhou2018sfcn,huang2020bcdata,abousamra2021multi,cai2021generalizing,zhang2022weakly,ryu2023ocelot}. The goal of PCD is to predict a 2D point set that represents the coordinates and classes of cells present in an input image. To accomplish this, prevalent PCD methods connect to off-the-shelf segmentation models via carefully crafted pseudo mask labels derived from point annotations. Subsequently, a series of post-processing steps comprising thresholding, local maxima detection and non-maximum suppression are applied to the predicted density maps to locate cells. However, the heuristic post-processing not only demands tedious hyper-parameter tuning but also results in low throughput. To address these issues, a recent study \cite{shui2022end} introduced Point-to-Point Network (P2PNet) \cite{song2021rethinking} to establish an end-to-end PCD system. Specifically, P2PNet adopts a detection paradigm, where the cell coordinates and categories can be directly obtained by refining and classifying pre-defined point proposals on an input image. Moreover, P2PNet employs a one-to-one matching scheme to suppress duplicate predictions, eliminating the need for error-prone and time-consuming post-processing. Because of these improvements, P2PNet can achieve superior accuracy and efficiency over the mainstream density map-based PCD methods.

Despite its successful application, we contend that the performance and flexibility of P2PNet can be limited from the following two aspects. (i) Unlike anchor boxes that naturally account for object scales, point proposals are scale-agnostic. As a result, P2PNet can only decode from a single-level feature map, which is inadequate to represent multi-scale information. Considering the substantial variability in cell size and morphology, as well as the heterogeneity in intensity distribution, decoding from hierarchical feature maps becomes imperative. (ii) The spatial distribution of artificially placed point proposals is inherently sub-optimal, with many of them situated far from cell centroids. This constitutes a significant challenge for achieving high-quality localization. On the other hand, P2PNet lacks the ability to perceive the positions of point proposals, which restricts it to use a fixed set of point proposals to ensure model convergence. To lift these limitations, this study proposes \textbf{\emph{d}}eformable \textbf{\emph{p}}roposal-\textbf{\emph{a}}ware P2PNet, dubbed as DPA-P2PNet. Overall, we incorporate two improvements into the vanilla P2PNet. First, we straightforwardly extract multi-scale decoding features for each point proposal according to its coordinates on feature pyramid. This modification not only enhances the model's capability but also makes it to be proposal-aware. Based on this, we further design deformable point proposals to reduce the distribution bias between proposals and potential cells to improve the localization quality and categorical discriminability of extracted features.
 
In clinical practice, pathologists generally perform accurate cell classification in two steps. They first zoom out to comprehend broad tissue structures and then zoom in to classify cells based on their morphology and the surrounding context. However, most computer-assisted PCD methods operate with a single image as input, deviating from the authentic diagnostic procedure. To mitigate this inconsistency, \cite{bai2020multi,bai2022context,ryu2023ocelot} adapt cell segmentation models to incorporate input of multi-field-of-view (mFoV) images, resulting in heightened  accuracy of density map-based PCD methods. Nonetheless, their improvements cannot be readily applied to end-to-end PCD models, owing to the intrinsic disparities in model architecture and operating mechanism. This study presents mFoV DPA-P2PNet as the first end-to-end PCD model capable of utilizing mFoV images for better cell classification. 

Self-supervised learning (SSL) aims to learn a generic representation applicable to various downstream tasks. Recent researches \cite{wang2021transpath,li2023task,kang2023benchmarking} have demonstrated that domain-aligned pre-training outperforms traditional transfer learning from ImageNet in several medical imaging tasks. Yet, the applicability of self-supervised domain-aligned pre-training to the dense prediction task of PCD has never been explored. To remedy this deficiency, we carry out an inaugural investigation into the effect of four representative SSL methods for the PCD task, including MoCo v2 \cite{chen2020improved}, SwAV \cite{caron2020unsupervised}, DINO \cite{caron2021emerging}, and MAE \cite{he2022masked}. Aside from the difference in the downstream task compared to prior studies, we are also the first to perform SSL on a large-scale and highly valuable immunohistochemistry (IHC) WSI dataset that involves three biomarkers, namely Ki-67, PD-L1, and HER-2.

Our main contributions can be summarized as follows:
\begin{itemize}
	\item We propose DPA-P2PNet, which significantly enhances the performance of P2PNet through the integration of two improvements, multi-scale decoding and deformable point proposals.
	\item We present mFoV DPA-P2PNet, which is the first end-to-end PCD model that leverages mFoV images to improve cell detection quality.
	\item To the best of our knowledge, this is the first work that conducts self-supervised pre-training on large-scale IHC image data and explores the utility of various SSL methods on the PCD task. The pre-trained weights will be released to fuel a myriad of computational pathology tasks.
\end{itemize}

\section{Related Work}
\subsection{Crowd counting and localization}
Crowd counting aims to estimate the number of people in an image. The mainstream idea is to regress a pseudo density map generated based on point annotations \cite{lin2022boosting,liang2022focal}. The final crowd count is calculated by 2D integration over the estimated density map.

To tackle the issues of local inconsistency \cite{lian2019density} and weak interpretability of density map-based crowd counting methods, recent studies \cite{abousamra2021localization,wan2021generalized, song2021rethinking,liang2022end,liang2022focal,lin2023optimal} have redirected their focus towards the localization-based crowd counting problem, where the crowd count is represented by the number of localized human heads. Existing crowd localization methods can be categorized into two types. The first line of methods conducts additional post-processing on the estimated density maps to localize individual human heads. Another type of methods comprising P2PNet \cite{song2021rethinking} and CLTR \cite{liang2022end} achieve end-to-end crowd localization by directly regressing the point coordinates. Currently, P2PNet represents the state-of-the-art in the field of crowd localization \cite{lin2023optimal}.

While crowd localization and PCD share similar high level spirit, there are two concretized differences between them. Firstly, in terms of localization, human heads have relatively regular and consistent shapes, whereas cells typically exhibit a wide range of shapes and sizes. This large variation poses a greater challenge for accurate cell localization. Secondly, when it comes to classification, crowd localization only concerns binary classes, distinguishing between foreground (heads) and background, while PCD generally involves multiple categories. Furthermore, classifying cells is significantly more demanding as it requires the integration of both coarse-grained tissue structure and fine-grained cell morphology. These two factors stress the necessity of leveraging multi-scale information for accurate cell detection.

\subsection{Point-based Cell Detection}
PCD aims to localize and classify cells in a pathology image, with each cell represented by a class-aware point. Mainstream PCD methods \cite{abousamra2021multi,cai2021generalizing,zhang2022weakly,ryu2023ocelot} operate similarly with density map-based crowd localization approaches but regress multiple density maps, each corresponding to a distinct cell type. Recently, \cite{shui2022end} introduces the advanced P2PNet to perform PCD in an end-to-end manner. However, the original P2PNet model can only decode from a single-level feature map, which is insufficient to squeeze the most out of the multi-scale information. To alleviate this deficiency, \cite{shui2022end} downsamples the shallow feature maps and executes feature fusion by element-wise summation, resulting in an enhanced feature map to decode. Nevertheless, we argue that this approach would lead to a loss of fine-grained information to some extent and the improved P2PNet still suffers from the biased distribution of pre-defined point proposals.

\subsection{Leveraging Large Field of Views}
Several studies \cite{kamnitsas2017efficient,tokunaga2019adaptive,ho2021deep,schmitz2021multi,van2021hooknet} extract a large FoV region as an additional input to improve the segmentation performance on smaller FoV regions. In the area of PCD, \cite{bai2020multi} and \cite{bai2022context} have devoted pioneering efforts in this direction. They propose a feature aggregation module that combines visual representations extracted from two images with different FoVs to enhance cell detection, where the tissue structure information is learned implicitly. A recent study \cite{ryu2023ocelot} incorporates the contextual knowledge explicitly via multi-task objectives at different FoVs. Specifically, they build two models for tissue and cell segmentation at large and small FoVs, respectively. The predicted tissue probability map is blended into the cell detection branch to promote cell classification. In this study, we adopt the same experimental setup as in \cite{bai2020multi,bai2022context} since the costly tissue mask annotation is unavailable in general. However, these two pioneering methods are both based on density map regression. So their improvements cannot be seamlessly applied to the advanced end-to-end PCD models due to the inherent discrepancy in model architecture and operating mechanism.

\subsection{Self-supervised Learning in Medical Imaging}
SSL has proven to be an effective method to learn a good representation from vast unlabeled images by solving a pre-text task. Pre-trained models from SSL are widely used in fine-tuning downstream tasks faster or for better accuracy \cite{chen2020adversarial}. For applications in medical imaging, transfer from ImageNet has become the de-facto approach. Yet, \cite{matsoukas2022makes} observes that domain-specific SSL methods can further improve the performance of models fine-tuned on downstream medical image-related tasks. This has been confirmed on numerous medical image analysis tasks such as pathology image classification \cite{sowrirajan2021moco,wang2021transpath,kang2023benchmarking,chen2022scaling,li2023task} and retrieval \cite{gildenblat2019self}, survival outcome prediction \cite{chen2022scaling}, nuclei instance segmentation \cite{kang2023benchmarking} and MRI brain tumor segmentation \cite{zhou2022self}. However, the applicability of self-supervised domain-aligned pre-training for the PCD task remains unrevealed.

\section{Approach}
In this section, we first briefly review the vanilla P2PNet. Then we elaborate the proposed DPA-P2PNet, including the multi-scale decoding (MSD) strategy and the generation of deformable point proposals (DPP). After that, we present mFoV DPA-P2PNet that supports mFoV images as input for better cell classification. Lastly, we detail the IHC dataset collected for SSL pre-training.

\subsection{Revisiting P2PNet}
Similar with modern object detectors \cite{ren2015faster}, P2PNet comprises three parts: backbone, neck and heads. Taken an image $I\in \mathbb{R}^{H\times W\times 3}$ as input, the backbone and neck produces hierarchical visual representations $\{P_i\}_{i=2}^L$. Let $s$ denotes the downsampling ratio (i.e., $s=2^i$), then the resolution of $P_i$ is $\frac{H}{s}\times\frac{W}{s}\times C$.

Since point proposals are scale-agnostic, P2PNet only selects one feature map (e.g., $P_i$) for decoding. Each location of $P_i$ corresponds to a $s\times s$ patch of the input image. To densely detect the cells, $m\times n$ point proposals are placed at each patch in a grid distribution, where $m$ and $n$ separately represent the number of rows and columns of point proposals. Conditional on $P_i$, two task-specific convolutional heads are employed to generate regression offset map and classification logit map of channels $2mn$ and $(C+1)mn$ while retaining the spatial resolution. $C$ is the number of cell types and the extra class is background. The 2D offsets and logits at each location are assigned to the point proposals within the corresponding patch.

\begin{figure}[t!]
	\begin{center}
		\includegraphics[width=0.98\linewidth]{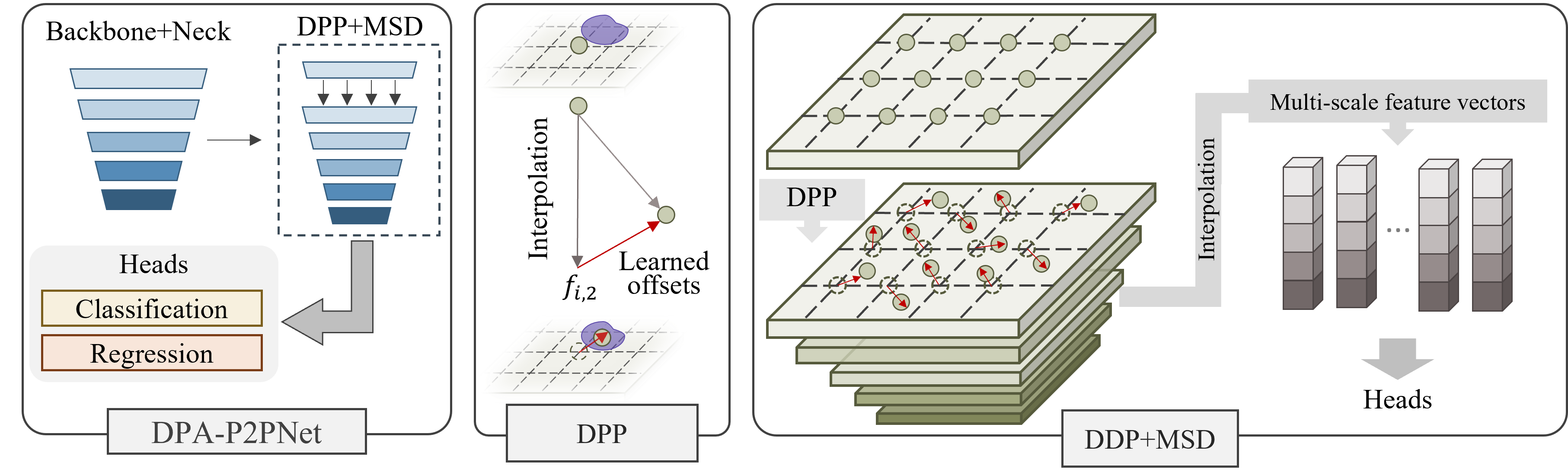}
		\caption{Framework of DPA-P2PNet.}
		\label{fig:framework}
		\vspace{-15pt}
	\end{center}
\end{figure}

\subsection{DPA-P2PNet}
Fig.~\ref{fig:framework} presents the overall framework of our proposed DPA-P2PNet. In the following sections, we use the set $\mathcal{S}=\{p_i\}_{i=1}^M$ to denote the pre-defined point proposals.

\paragraph{Decoding from feature pyramid}
To effectively exploit the information at different granularities, we construct ROI features from hierarchical feature maps based on the coordinates of each point proposal. Specifically, the ROI feature vectors $\{f_{i,j}\}_{j=1}^{L}$ for proposal $p_i$ are extracted from the feature pyramid via the bilinear interpolation method:
\begin{equation}
	{f_{i,j}} = \sum_{q}G(p_i, q)\cdot P_j(q),
\end{equation}
where $j$ denotes the feature level, $q$ enumerates all integral spatial locations around $p_i$ in the feature map $P_j$ and $G(\cdot,\cdot)$ is the bilinear interpolation kernel. Then, we concatenate $\{f_{i,j}\}_{j=1}^{L}$ and fed it into two dedicated MLP heads for decoding offsets and logits with respect to $p_i$.

The decoding strategy described above not only facilitates the exploitation of multi-scale information but also endows our model with a proposal-aware ability. Intuitively, in the original P2PNet, a point proposal passively receives the decoded content, whereas in our model, proposals actively query the distances to potential cells. As a result, P2PNet is confined to a fixed set of point proposals to ensure model convergence. However, a considerable portion of these manually placed proposals deviate far from cell centroids, which could easily lead to inaccurate localization. This distribution bias can also cause the extracted features to be less discriminative for cell classification in our model. The proposal-aware nature opens up possibilities for dynamically refining the point proposals without concerns of model dispersion.

\begin{figure}[t!]
	\begin{center}
		\includegraphics[width=0.98\linewidth]{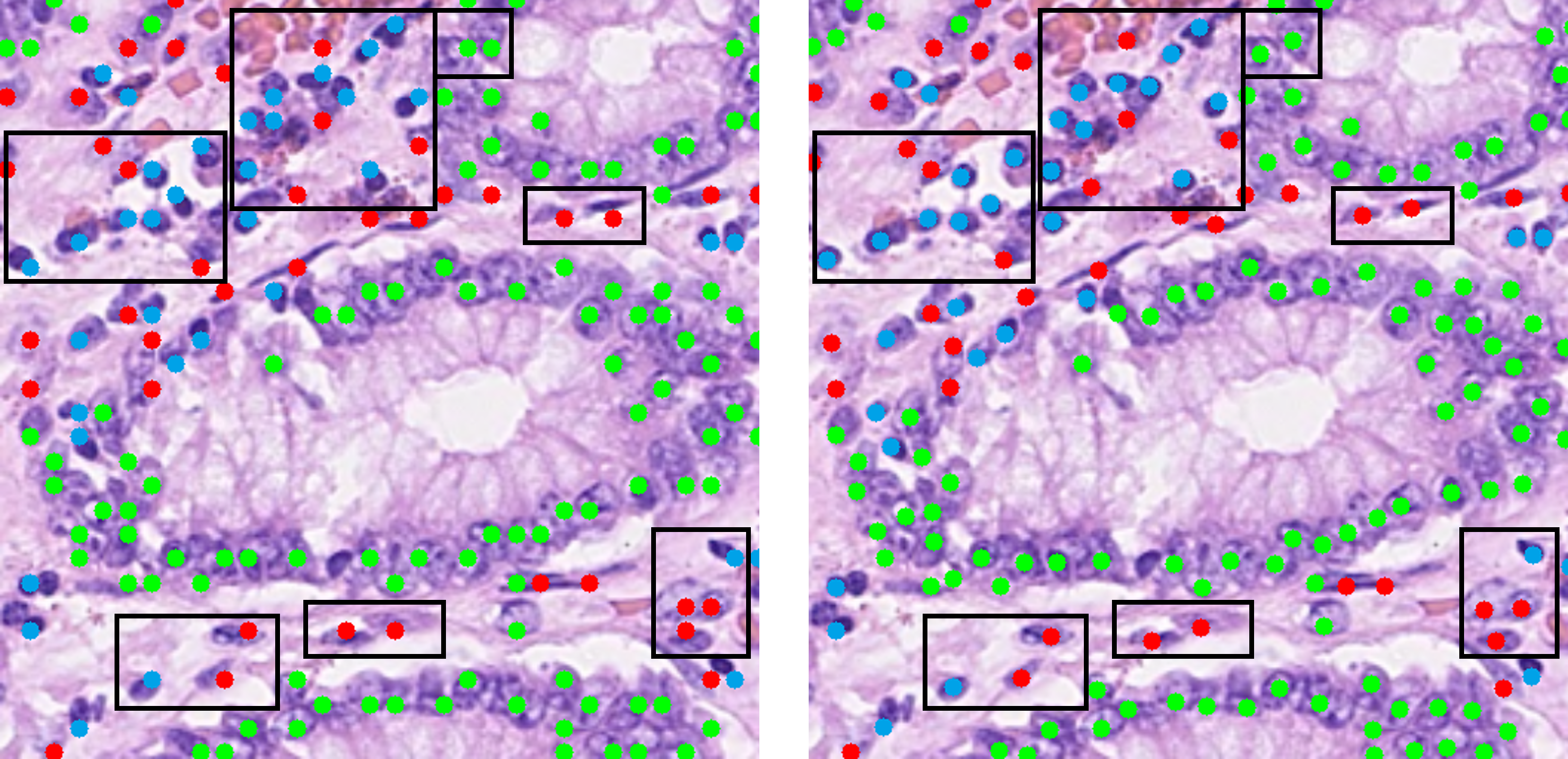}
		\caption{Illustration of our proposed deformable point proposals. The positions of foreground proposals before (left) and after (right) the deformation are depicted. We highlight some ROI regions with boxes for comparison. For a clearer view, we encourage readers to zoom in. \textcolor{red}{$\mdlgblkcircle$} Epithelial Cells, \textcolor{green}{$\mdlgblkcircle$} Stromal Cells, \textcolor{lightblut}{$\mdlgblkcircle$} Inflammatory Cells.}
		\label{fig:deform}
		\vspace{-20pt}
	\end{center}
\end{figure}

\paragraph{Deformable point proposals}
Improving the quality of object proposals has been a long-standing research topic in the field of object detection. Recent one-stage detectors \cite{zhu2020deformable,sun2021sparse,zhang2022dino} commonly adopt an iterative refinement strategy, which prompts the current head to refine the locations of bounding boxes predicted by the previous head and ultimately gathers all box proposals into the label assignment module. However, in the case of one-to-one matching, this approach exacerbates the foreground-background imbalance since the number of positive proposals remains constant while the total number of proposals increase. This imbalance could potentially lead to unexpected performance degradation. To address this issue, we propose a structurally simple yet highly potent solution to enhance the quality of point proposals without increasing their quantity.

As illustrated in Fig.~\ref{fig:framework}, we deform the pre-set point proposals to reduce their distances to neighboring cells. To achieve this, we use a MLP layer to generate the deformation offsets from $f_{i,2}$, informed by that the high-resolution $P_2$ contains the finest-grained features essential for small object localization \cite{lin2017feature}. It is noteworthy that the offsets are learned implicitly without direct supervision. Based on the deformed point proposals $\mathcal{S}^\prime$, we proceed with the decoding procedure, as described in the preceding section, to perform classification and finer localization.

Fig.~\ref{fig:deform} presents an intuitive demonstration of the deformation process. Clearly, the pre-defined point proposals are adaptively transported onto nearby cells by this transformation. On the basis of $\mathcal{S}^\prime$, we can extract more discriminative features for classification. Moreover, through additional refinement during the subsequent decoding stage, the localization quality can be further improved.

\subsection{mFoV DPA-P2PNet}
Multiple concentric images $\{I^k\}_{k=1}^K$ with the same resolution yet captured at different objective magnification can be interpreted as having multiple FoVs. In this paper, we use $I^K$ to represent the annotated image with the highest magnification yet smallest FoV, while $\{I^k\}_{k=1}^{K-1}$ to denote the set of  unlabeled images with larger FoVs. Following the setup of \cite{bai2020multi,bai2022context}, the magnification of $I^k$ is twice that of $I^{k-1}$.

\begin{figure}[t!]
	\begin{center}
		\includegraphics[width=0.98\linewidth]{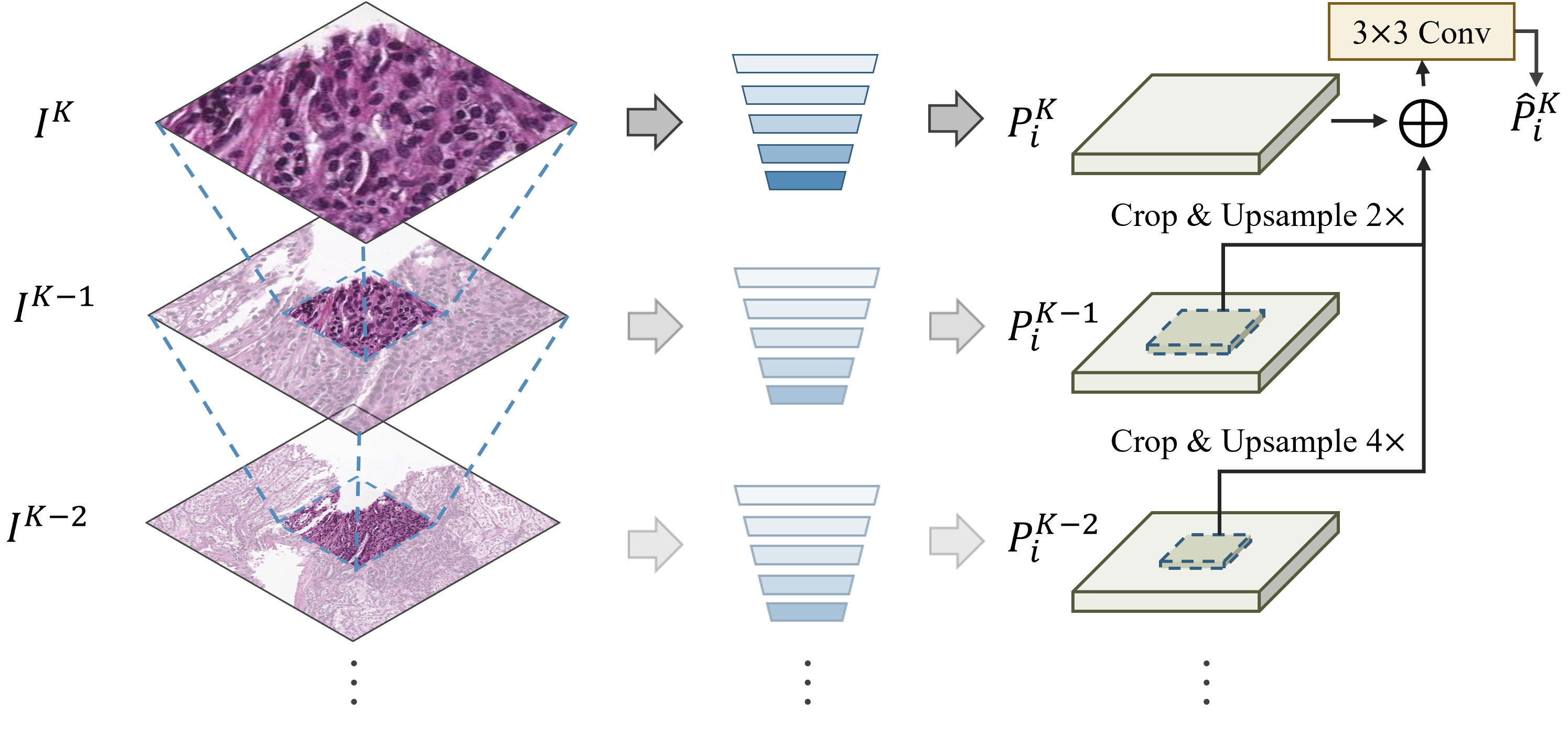}
		\caption{Schematic of mFoV DPA-P2PNet.}
		\label{fig:mfov}
		\vspace{-10pt}
	\end{center}
\end{figure}

\begin{table}[t!]
	\centering
	\small{
		\resizebox{0.99\linewidth}{!}{
			\begin{tabular}{c|c|c|c}
				\shline
				IHC biomarker & No. of WSIs & No. of patches & Cancer type \\
				\shline
				Ki-67 & 403 & 233,308 & breast and stomach cancers \\
				PD-L1 & 1,208 & 607,302 & non-small cell lung cancer \\
				HER2 & 1,101 & 597,563 & breast cancer \\
				\shline 	
	\end{tabular}}}
	\caption{IHC data for pre-training.}
	\vspace{-15pt}
	\label{tab:pretraining_dataset}
\end{table}

\begin{table*}[t!]
	\centering
	\small{
		\resizebox{0.99\linewidth}{!}{
			\begin{tabular}{c|c|ccc|ccc|c|c|c}
				\shline
				\multirow{2}{*}{Datasets} & Objective & 
				\multicolumn{3}{c|}{No. of patches} & \multicolumn{3}{c|}{No. of cells} &
				No. of & \multirow{2}{*}{Image resolution} & \multirow{2}{*}{Organs} \\
				\cline{3-8}
				& magnification & train & val & test & train & val & test & categories & & \\
				\shline
				CoNSeP & 20$\times$ & 22 & 5 & 14 & 13,040 & 2,515 & 8,777 & 3 & 500$\times$500 & colon \\
				BCData & 40$\times$ & 803 & 133 & 402 & 93,838 & 21,804 & 65,432 & 2 & 640$\times$640 & breast \\
				PD-L1 & 40$\times$& 1215 & 405 & 405 & 358,832 & 116,148 & 116,418 & 10 & 1024$\times$1024 & lung \\
				OCELOT & 30$\times$& 252 & 78 & 70 & 41,645 & 13,574 & 10,618 & 2 & 1024$\times$1024 & kidney, stomach, etc. \\
				\shline
	\end{tabular}}}
	\caption{Profiles of four histopathology datasets. A 20$\times$ objective magnification corresponds to approximately 0.5 $\mu$m/px.}
	\label{tab:dataset}
	\vspace{-10pt}
\end{table*}

Fig.~\ref{fig:mfov} presents the framework of mFoV DPA-P2PNet. Specifically, we first use a separate backbone and neck to construct the feature pyramid $\{P_i^k\}_{i=2}^L$ for each image. After that, we aggregate the contextual information extracted from large FoV images into the feature maps $\{P_i^K\}_{i=2}^L$ obtained from $I^K$, which can be expressed as: 
\begin{equation} \label{eq:mfov}
	\hat{P}_i^K={\rm Conv}\left(P_i^K+\sum_{k=1}^{K-1}{\rm Upsample}\left({\rm Crop}\left(P_i^k\right)\right)\right)
\end{equation} where ${\rm Crop}(\cdot)$ is the center square crop operation. For $P_i^k$, the cropping limit normalized by dividing the height or width is $\left[\frac{2^{K-k}-1}{2^{K-k+1}},\frac{2^{K-k}+1}{2^{K-k+1}}\right)$. Then we upsample the cropped region with a factor of $2(K-k)$ to match the resolution of $P_i^K$ and fuse the mFoV features by element-wise summation. Unless specified otherwise, the upsampling process is completed by a shared ConvTranspose2d layer in this work. Thereafter, in order to eliminate the aliasing effect of upsampling \cite{lin2017feature}, we employ a 3$\times$3 convolution on the merged map to generate the final feature map $\hat{P}_i^K$. Finally, we utilize the enhanced feature pyramid $\{\hat{P}_i^K\}_{i=2}^L$ for decoding.

\subsection{Data Collection for SSL}
Tab.~\ref{tab:pretraining_dataset} provides an overview of the immunohistochemistry (IHC) dataset used for pre-training. We first collect 2,712 WSIs that cover three types of IHC biomarkers: Ki-67, programmed death-ligand 1 (PD-L1), and human epidermal growth factor receptor 2 (HER2). To increase the diversity and informativeness of the pre-training dataset, we extract at most 500 patches of resolution 512 $\times$ 512 pixels and objective magnification 40$\times$ (0.25 $\mu$m/px) from each WSI. Moreover, a pre-trained cell detector is deployed to guarantee that each preserved patch contains at least 30 cells. As a result, we collect a total of 1.4M patches, slightly exceeding the scale of the ImageNet-1K training set \cite{deng2009imagenet}.

\paragraph{Broader impact} Beyond assessing the effect of various SSL algorithms on the PCD task, we are the first to conduct self-supervised pre-training with large-scale IHC image data. In contrast to HE data that is widely adopted in previous relevant works, IHC staining allows the detection and localization of specific proteins or antigens within tissue sections, which is particularly valuable for identifying specific cell types, biomarkers, or pathological changes associated with diseases. In clinical practice, pathologists usually combine the complementary merits of HE and IHC stained WSIs to obtain a more comprehensive understanding of tissue samples. By making the pre-trained weights publicly available, we hope to advance the development of the computational pathology community.

\section{Experiment}

\begin{table*}[t!]
	\centering
	\resizebox{\linewidth}{!}{	
		\begin{tabular}{c|c|c c c c c c c c | c  c c c}
			\toprule[1.5pt]
			\multirow{2}{*}{Datasets} & \multirow{2}{*}{Metrics} &
			\multicolumn{8}{c|}{\cellcolor{gray!40}\emph{DM-based Methods}} & \multicolumn{4}{c}{\cellcolor{gray!40}\emph{End-to-end Methods}} \\ \cline{3-14}
			& & \cellcolor{gray!40}U-Net & \cellcolor{gray!40}DeepLabV3+ & \cellcolor{gray!40}U-CSRNet & \cellcolor{gray!40}MCSpatNet & \cellcolor{gray!40}ML-CellNet & \cellcolor{gray!40}GL & \cellcolor{gray!40}FIDT & \cellcolor{gray!40}OT-M & \cellcolor{gray!40}P2PNet & \cellcolor{gray!40}CLTR &  \cellcolor{gray!40}E2E &
			\cellcolor{gray!40}Ours \\ \bottomrule
			\toprule			
			\multirow{2}{*}{CoNSeP} & F1 & 61.8 & 65.7 & 42.5 & 68.2* & 37.1 & 16.7 & \underline{70.3} & 36.7 & 70.0 & 40.8 & 70.2 & \textbf{71.1} \\
			& AP & - & 53.7 & 27.7 & 52.2* & 20.0 & - & 57.7 & - & \underline{60.5} & 21.8 & 59.9 & \textbf{62.9} \\ \hline
			\multirow{2}{*}{BCData} & F1 & 85.7 & 85.8 & 85.7* & 85.3 & 84.6 & 68.5 & 85.9 & 66.5 & \underline{86.3} & 84.8 & \underline{86.3} & \textbf{86.7} \\			
			& AP & 80.1 & 82.1 & - & 79.2 & 79.7 & - & 81.3 & - & \underline{83.8} & 81.0 & 83.3 & \textbf{84.4} \\ \hline
			\multirow{2}{*}{PD-L1} & F1 & 46.0 & 53.7 & 25.7 & 43.6 & 30.0 & 32.5 & 51.9 & 27.1 & 54.4 & 54.3 & \underline{54.8} & \textbf{55.9} \\
			& AP & 33.4 & 41.3 & 21.1 & 30.8 & 20.7 & - & 36.4 & - & 42.1 & 43.0 & \underline{42.5} & \textbf{43.7} \\ \hline
			\multirow{3}{*}{PD-L1} & Params(M) & 31.0 & 40.3 & 10.1 & 26.1 & 40.3 & 21.5 & 66.6 & 21.5 & 27.3 & 41.0 & 28.9 & 32.3 \\
			& MACs(G) & 876 & 277 & 184 & 1371 & 277 & 432 & 579 & 432 & 100 & 105 & 107 & 151 \\
			& FPS & 14 & 28 & 3 & 12 & 27 & 27 & 10 & 2 & 41 & 22 & 42 & 39 \\
			\bottomrule[1.5pt]
		\end{tabular}
	}
	\caption{\label{tab:performance} Quantitative comparison on three datasets. * indicates the previously publicly reported best results on the dataset. The best and second-best performance are highlighted in \textbf{bold} and \underline{underlined}, respectively.}
	\vspace{-10pt}
\end{table*}

\subsection{Experimental Setup}
\subsubsection{Dataset}
In this study, we evaluate the advantages of DPA-P2PNet over the state-of-the-art counterparts on three histopathology datasets with varied staining types, including the HE stained CoNSeP \cite{graham2019hover}, IHC Ki-67 stained BCData \cite{huang2020bcdata} datasets, and an internal IHC PD-L1 dataset. To validate the efficacy of our proposed mFoV DPA-P2PNet, we conduct comprehensive experiments on the OCELOT \cite{ryu2023ocelot} dataset, which offers the mapping from annotated patches to their source WSIs in TCGA \cite{hutter2018cancer} so that we can crop patches at arbitrary FoVs. We divide the PD-L1 and OCELOT dataset into \textit{training}, \textit{validation}, and \textit{test} subsets at a ratio of 6:2:2. To avoid information leaking among the subsets, we randomly split the dataset per WSI, ensuring that different patches from the same WSI are not included in multiple subsets. Tab.~\ref{tab:dataset} provides the statistics of these datasets, and detailed cell categories are available in the supplementary material.

\paragraph{Implementation Details}
The interval of pre-defined point proposals is set to 8 pixels on the CoNSeP dataset while 16 pixels on the other three datasets. By default, we use ResNet-50 \cite{he2016deep} and FPN \cite{lin2017feature} as the trunk and neck networks, respectively. All MLPs are structured as FC-ReLu-Dropout-FC. With the same label assignment scheme and loss functions as P2PNet \cite{song2021rethinking}, we adopt the AdamW optimizer \cite{loshchilov2017decoupled} with weight decay 1e-4 to optimize our proposed models. During the training stage, data augmentations including random scaling, shifting and flipping are applied on the fly. For the pre-training, we utilize the configurations recommended in \cite{kang2023benchmarking} to execute various SSL algorithms on our collected IHC dataset. All models are trained on NVIDIA A100 GPUs.

\subsubsection{Evaluation Metrics} We adopt macro-average F1-score and average precision (AP) as metrics to measure the cell detection performance of all models. Following \cite{ryu2023ocelot}, if a detected cell is within a valid distance ($\approx$3$\mu$m) from an annotated cell and the cell class matches, it is counted as a true positive (TP), otherwise a false positive (FP). Accordingly, considering the magnification of each dataset, we set the matching distance as 6, 12, 9 pixels for the CoNSeP, PD-L1 and OCELOT datasets, respectively. As for the BCData, we adhere to its official setup and use the threshold of 10 pixels to enable meaningful comparison.

\subsection{Experimental Results}
In the following parts, we first compare the capability of DPA-P2PNet with the state-of-the-art PCD and crowd localization competitors, which encompass density map (DM)-based approaches containing U-Net \cite{ronneberger2015u}, DeepLabV3+ \cite{chen2018encoder,ryu2023ocelot}, U-CSRNet \cite{huang2020bcdata}, MCSpatNet \cite{abousamra2021multi}, ML-CellNet \cite{zhang2022weakly}, GL \cite{wan2021generalized}, FIDT \cite{liang2022focal} and OT-M \cite{lin2023optimal}, as well as end-to-end methods including P2PNet \cite{song2021rethinking}, CLTR \cite{liang2022focal} and E2E \cite{shui2022end} on the CoNSeP, BCData and PD-L1 datasets. Subsequently, we demonstrate the superiority of mFoV DPA-P2PNet over the precursor studies \cite{bai2020multi,bai2022context} on the OCELOT dataset. Thereafter, we conduct several ablation experiments to show the effectiveness of MSD and DPP. Finally, we test the validity of diverse SSL methods on the PCD task.

\subsubsection{Comparison with SOTA Methods}
Table~\ref{tab:performance} shows the quantitative comparison results of our approach with the counterparts. Briefly, our proposed DPA-P2PNet achieves the highest F1 and AP scores on all datasets. The DM-based methods generally exhibit inferior performance because they rely on post-processing to localize cell centroids on the predicted density maps. However, finding a set of post-processing parameters that work well in all scenes is impossible. For example, setting the confidence threshold too large inevitably filters out cells with weak intensity, while using a small confidence threshold makes it difficult to separate overlapping cells. The inferior performance of P2PNet compared to our method can be attributed to its insufficient utilization of multi-scale information and the sub-optimal distribution of pre-defined point proposals. E2E unifies the resolutions of multi-level feature maps and performs feature fusion by element-wise summation, which mitigates the former drawback of P2PNet but somewhat leads to a loss of information. In comparison, DPA-P2PNet can fully utilize multi-scale features via  instant decoding from the uncompressed feature pyramid. Although the transformer-based CLTR model employs a more elegant query-based paradigm compared to our method, it requires a larger amount of labeled data to unleash its potential because vision transformer (ViT) based models have weaker inductive bias than CNNs in modeling visual structures and thus require much more labels to learn such bias implicitly \cite{xu2021vitae}. It can be observed from Tab.~\ref{tab:performance} that the performance gap between CLTR and DPA-P2PNet gradually decreases with the increase of training data. However, obtaining cell annotations on a large scale demands specialized expertise, and incurs a huge cost. In this context, our model is more label-efficient. Additionally, the capacity of CLTR, quantified by the number of queries, cannot scale with resolutions of input images, which greatly curtails its potential for clinical application.

\begin{table}[t!]
	\centering
	\resizebox{0.99\linewidth}{!}{	
		\begin{tabular}{c|c|c c }
			\toprule[1.5pt]
			& Methods & F1 & AP \\
			\bottomrule
			\toprule
			\multirow{3}{*}{DeepLabV3+} & Baseline & 58.8 & 42.7 \\
			& MFoVCE-Net & 59.3 & 43.5 \\		 	 
			& MFoVCE-Net+ & 59.8 & 44.2 \\ \hline
			\multirow{3}{*}{DPA-P2PNet} & Baseline & 59.3 & 44.4 \\			
			& Ours (bilinear interpolation) & \underline{61.9} & \underline{48.9} \\
			& Ours (transposed convolution) & \textbf{62.6} & \textbf{49.1} \\						
			\bottomrule[1.5pt]
		\end{tabular}
	}
	\caption{\label{tab:mfov} Performance comparison of various PCD methods with mFoV inputs. $K$ is set as 2. The baselines represent using only the small FoV patch.}
	\vspace{-15pt}
\end{table}

We also analyze the model size, computational cost and inference efficiency of different methods on the PD-L1 dataset in Tab.~\ref{tab:performance}. The DM-based methods demonstrate high computational complexity because they need to regress the high-resolution cell density maps. In contrast, the end-to-end models operate with hidden features of lower resolution and directly output cell coordinates and categories without the need for time-consuming post-processing. Consequently, they require fewer computational resources and generally exhibit faster inference speeds compared to the DM-based methods. The CLTR model, however, stands as an exception due to the considerable time complexity of the multi-head self-attention layers. Despite a slight sacrifice in speed, our proposed DPA-P2PNet outperforms the original P2PNet notably by 1.5\% on F1 and 1.6\% on AP. It is worth noting that the increased parameters and computational cost of our model over P2PNet is arguably fixed, which implies that the relative gaps would be inconsequential when adopting more powerful yet complex backbones or necks to enhance detection performance. For example, by replacing the ResNet-50 with ViT-Adapter-B/16 \cite{chen2022vision}, as practiced in Tab.~\ref{tab:ssl}, the relative gap is reduced from 18\% (32.3M vs. 27.3M) to 2\% (103.0M vs. 100.7M) on model size and from 41\% (151G vs. 100G) to 9\% (709G vs. 649G) on MACs.

\begin{table}[t]
	\begin{minipage}[t!]{0.5\linewidth}
		\centering
		\resizebox{1.0\linewidth}{!}{	
			\begin{tabular}{c|c c c}
				\toprule[1.5pt]
				No. of FoVs & F1 & AP & FPS \\
				\bottomrule
				\toprule
				1 & 59.3 & 44.4 & 25 \\
				2 & 62.6 & 49.1 & 18 \\
				3 & 64.2 & 51.4 & 11 \\
				4 & 64.6 & 51.6 & 8 \\
				\bottomrule[1.5pt]
			\end{tabular}
		}
		\caption{Cell detection performance of mFoV DPA-P2PNet on the OCELOT dataset as $K$ increases from 1 to 4.}
		\vspace{-15pt}
		\label{tab:num_fov}
	\end{minipage}
	\hfill
	\begin{minipage}[t!]{0.45\linewidth}
		\centering
		\resizebox{1.0\linewidth}{!}{	
			\begin{tabular}{c c|c c }
				\toprule[1.5pt]
				MSD & DPP & F1 & AP \\
				\bottomrule
				\toprule
				& & 70.0 & 60.5 \\
				\cmark & & 70.7 & 61.7 \\
				\cmark & \cmark & 71.1 & 62.9 \\
				\bottomrule[1.5pt]
			\end{tabular}	
		}
		\caption{Effect of MSD and DPP on the CoNSeP dataset. P2PNet serves as the baseline.}
		\vspace{-15pt}
		\label{tab:msd_dpp}
	\end{minipage}
\end{table}

\begin{figure*}[t!]
	\centering
	\begin{minipage}{0.16\textwidth}
		\includegraphics[width=\linewidth]{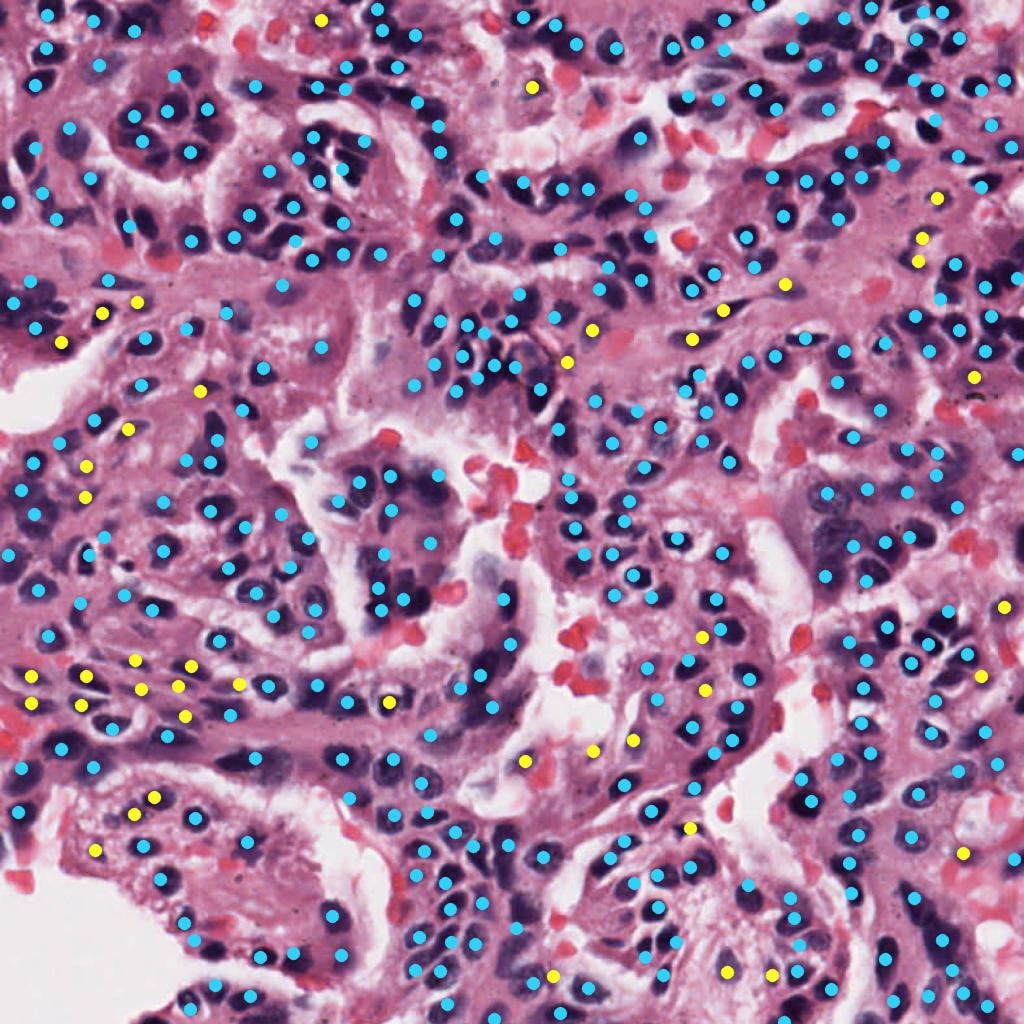}
	\end{minipage}
	\begin{minipage}{0.16\textwidth}
		\includegraphics[width=\linewidth]{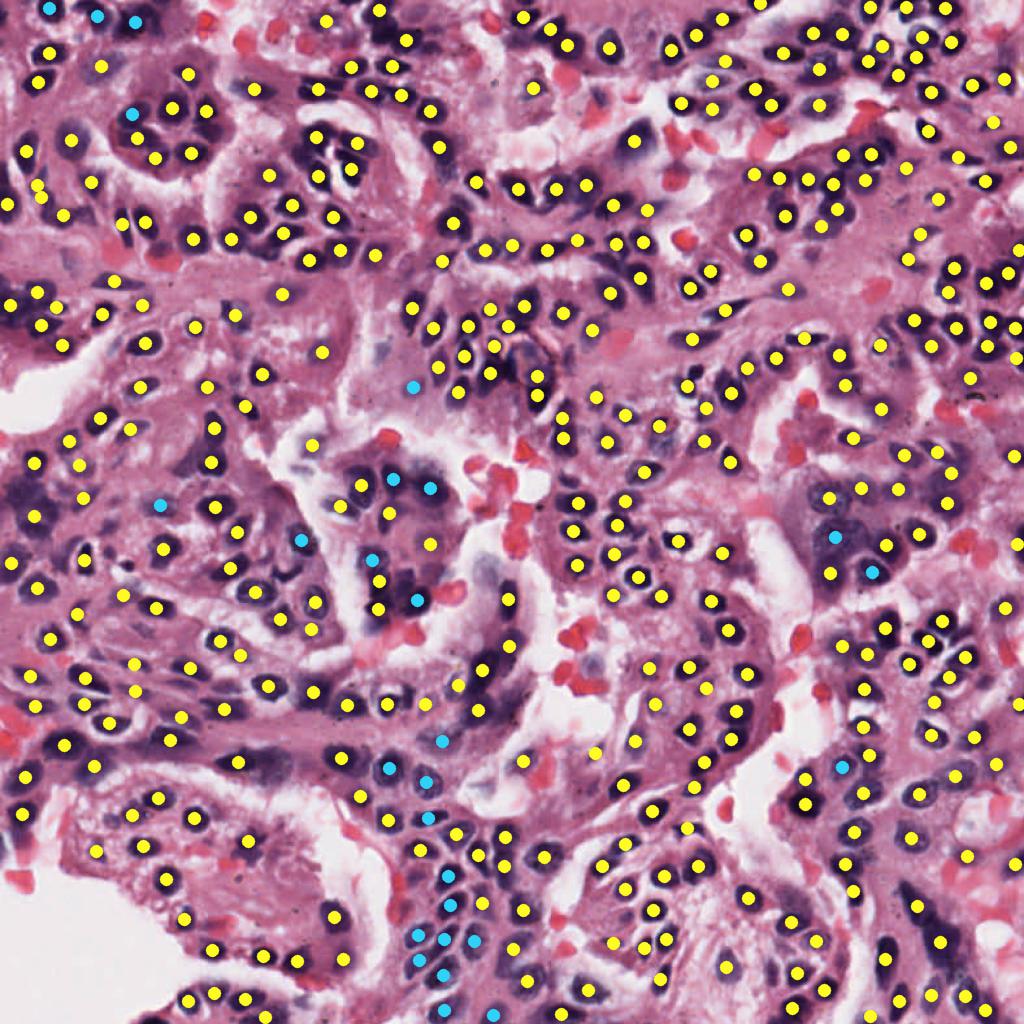}
	\end{minipage}
	\begin{minipage}{0.16\textwidth}
		\includegraphics[width=\linewidth]{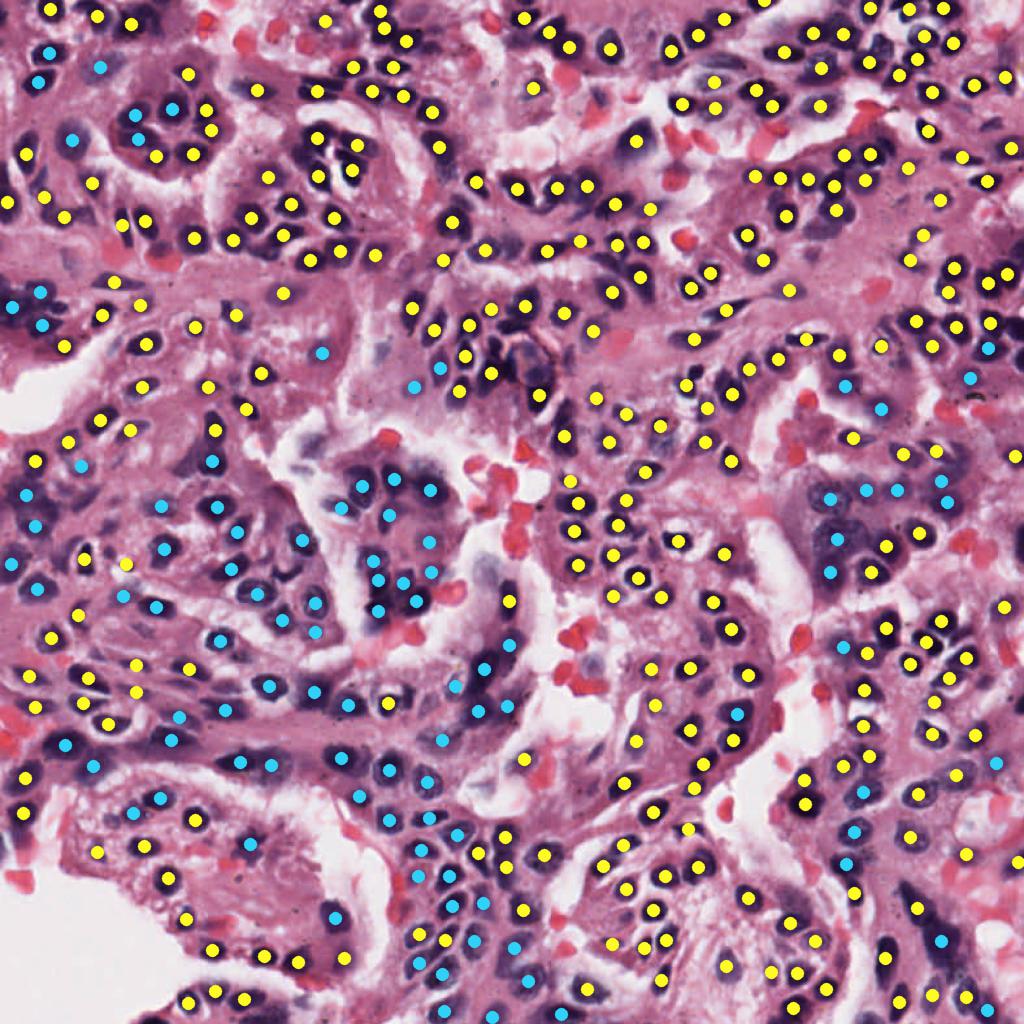}
	\end{minipage}
	\begin{minipage}{0.16\textwidth}
		\includegraphics[width=\linewidth]{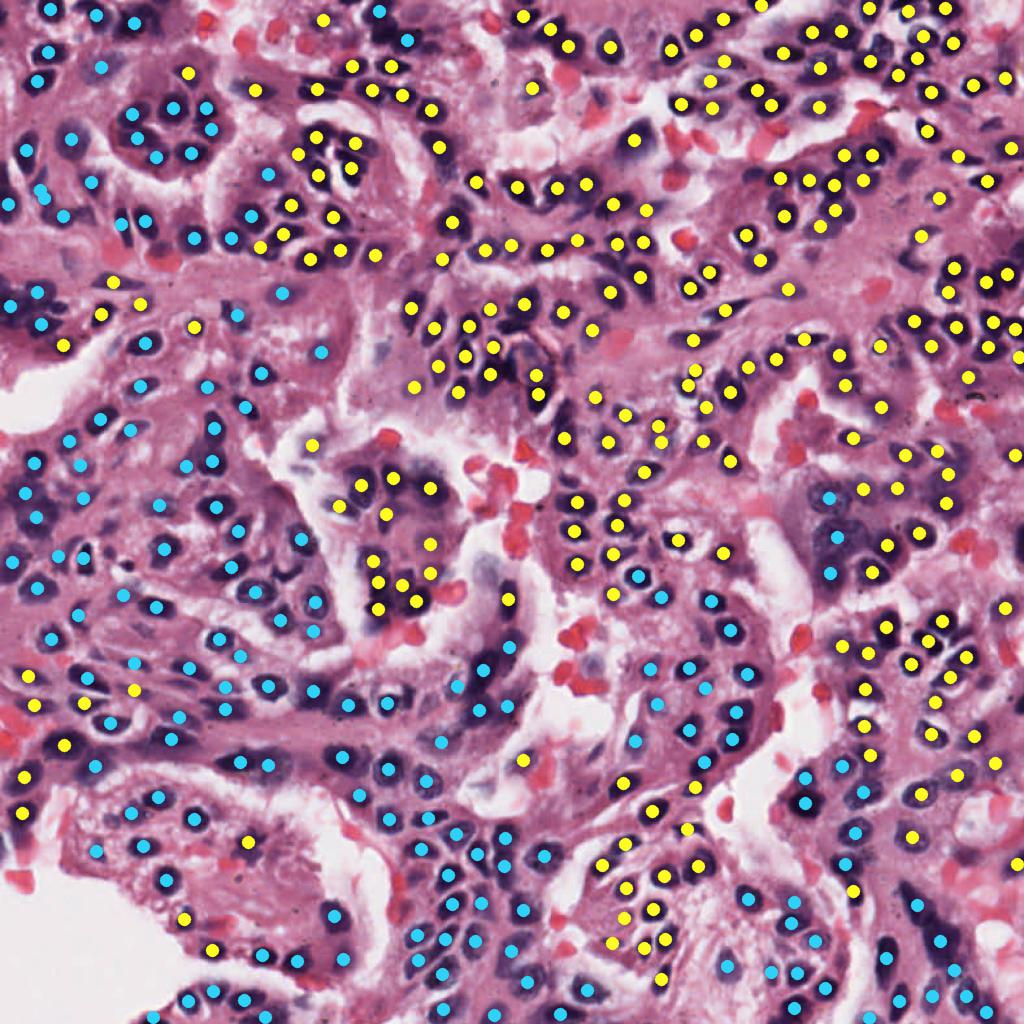}
	\end{minipage}
	\begin{minipage}{0.16\textwidth}
		\includegraphics[width=\linewidth]{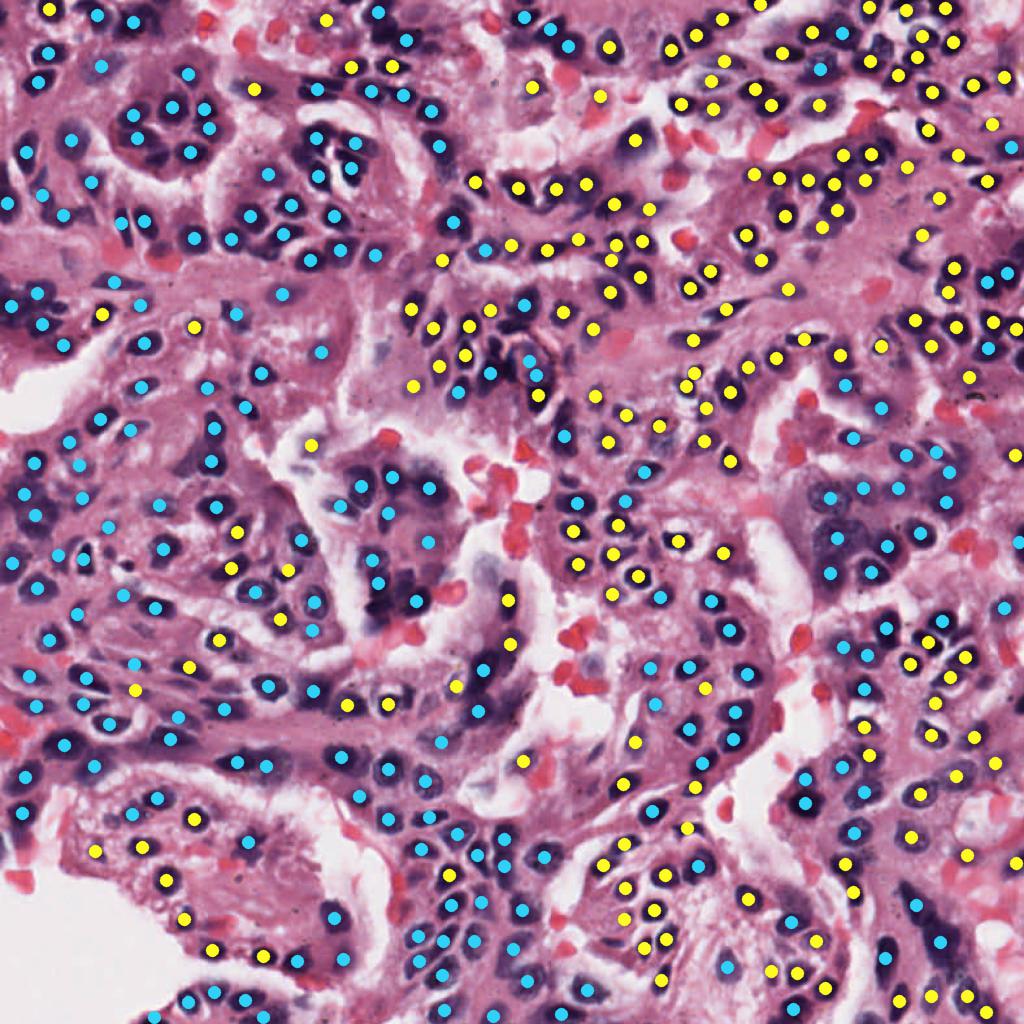}
	\end{minipage}
	\begin{minipage}{0.16\textwidth}
		\includegraphics[width=\linewidth]{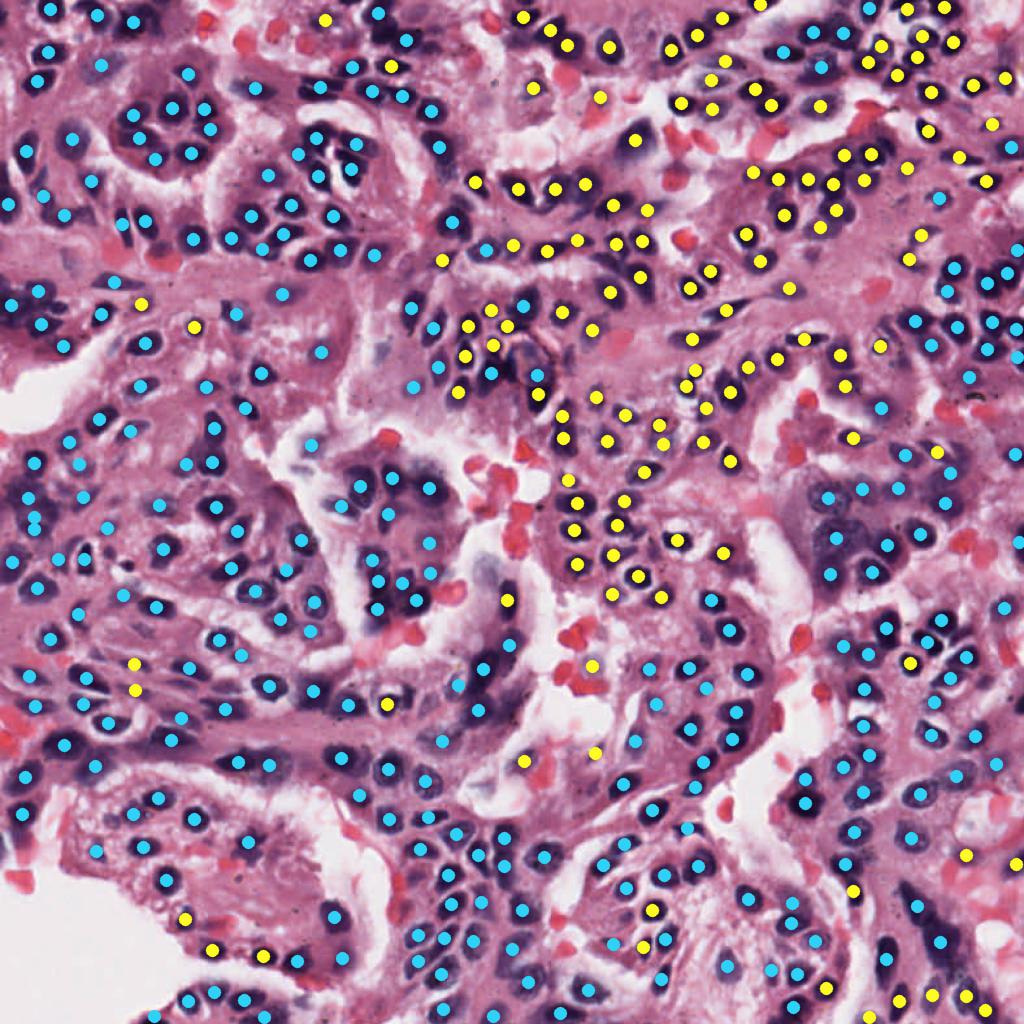}
	\end{minipage} \\
	\vspace{3pt}
	\begin{minipage}{0.16\textwidth}
		\includegraphics[width=\linewidth]{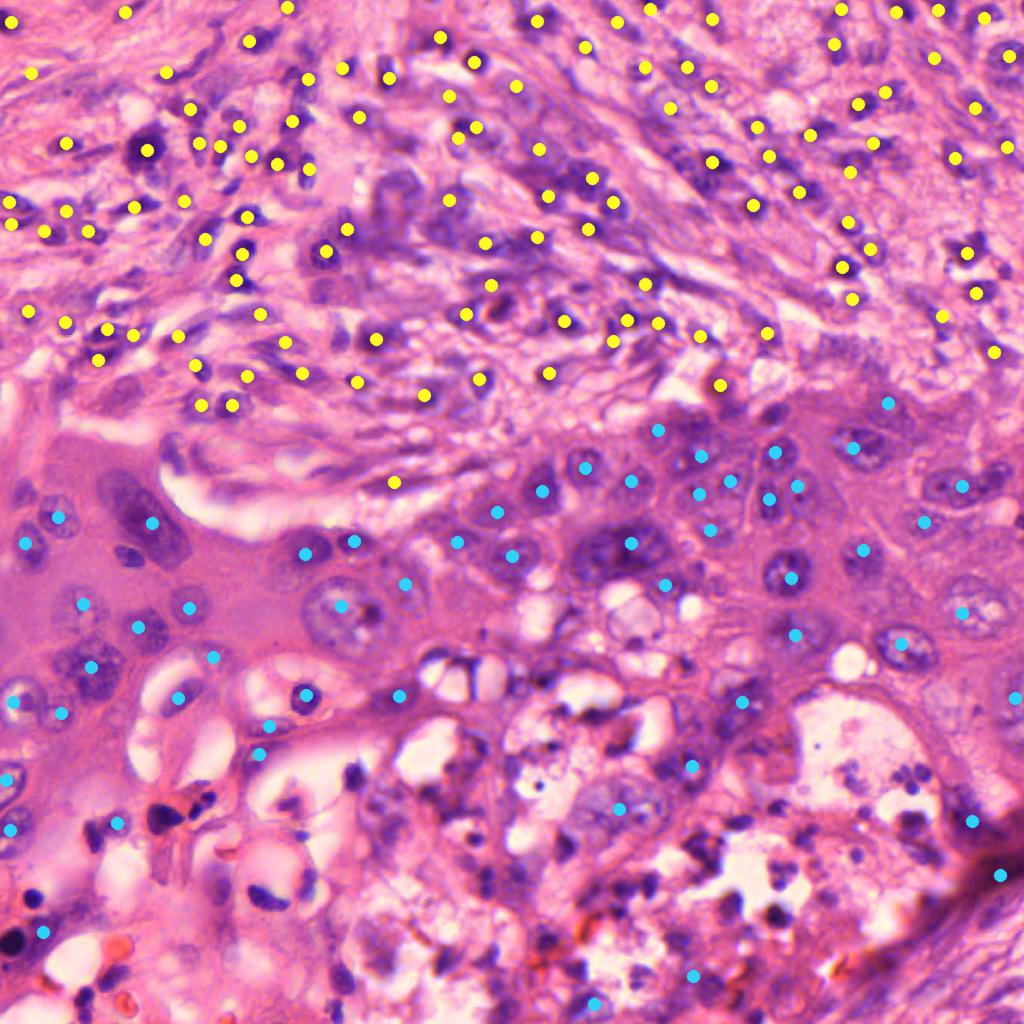}
	\end{minipage}
	\begin{minipage}{0.16\textwidth}
		\includegraphics[width=\linewidth]{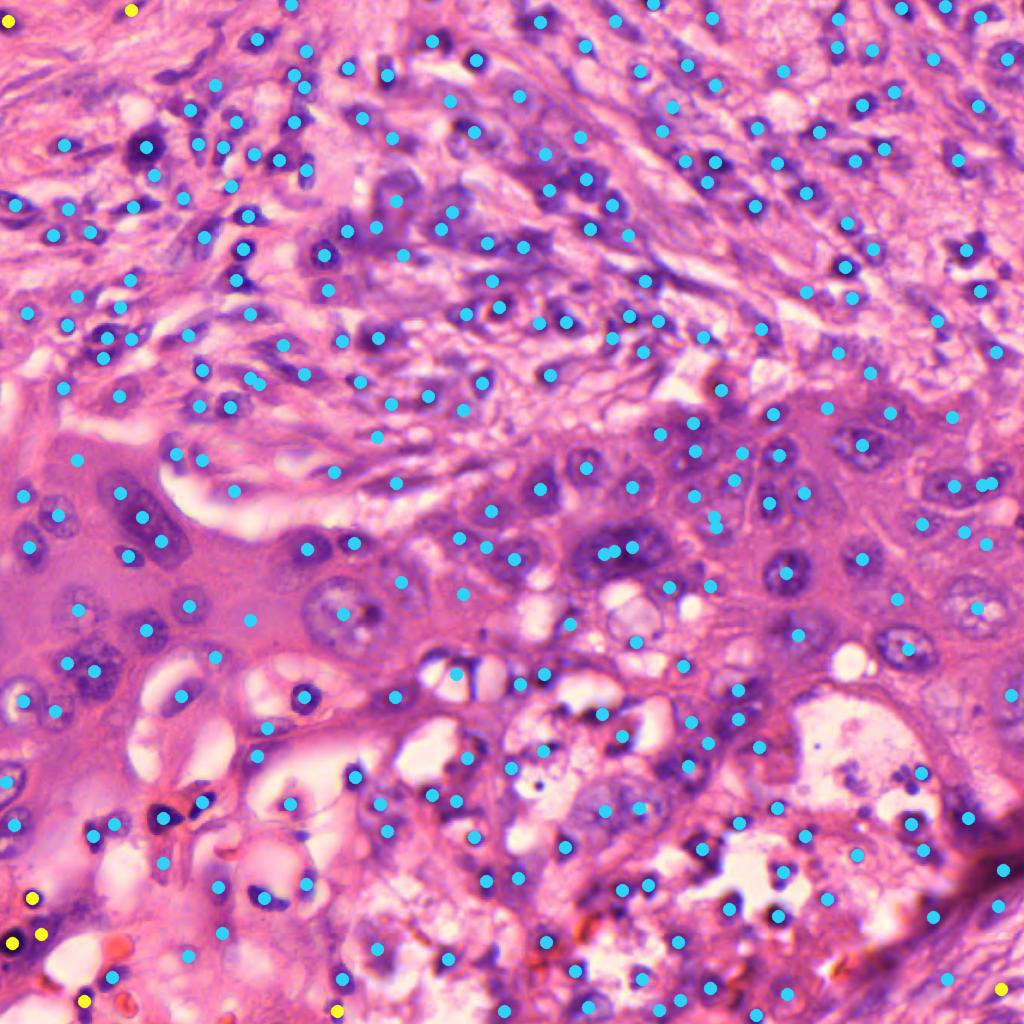}
	\end{minipage}
	\begin{minipage}{0.16\textwidth}
		\includegraphics[width=\linewidth]{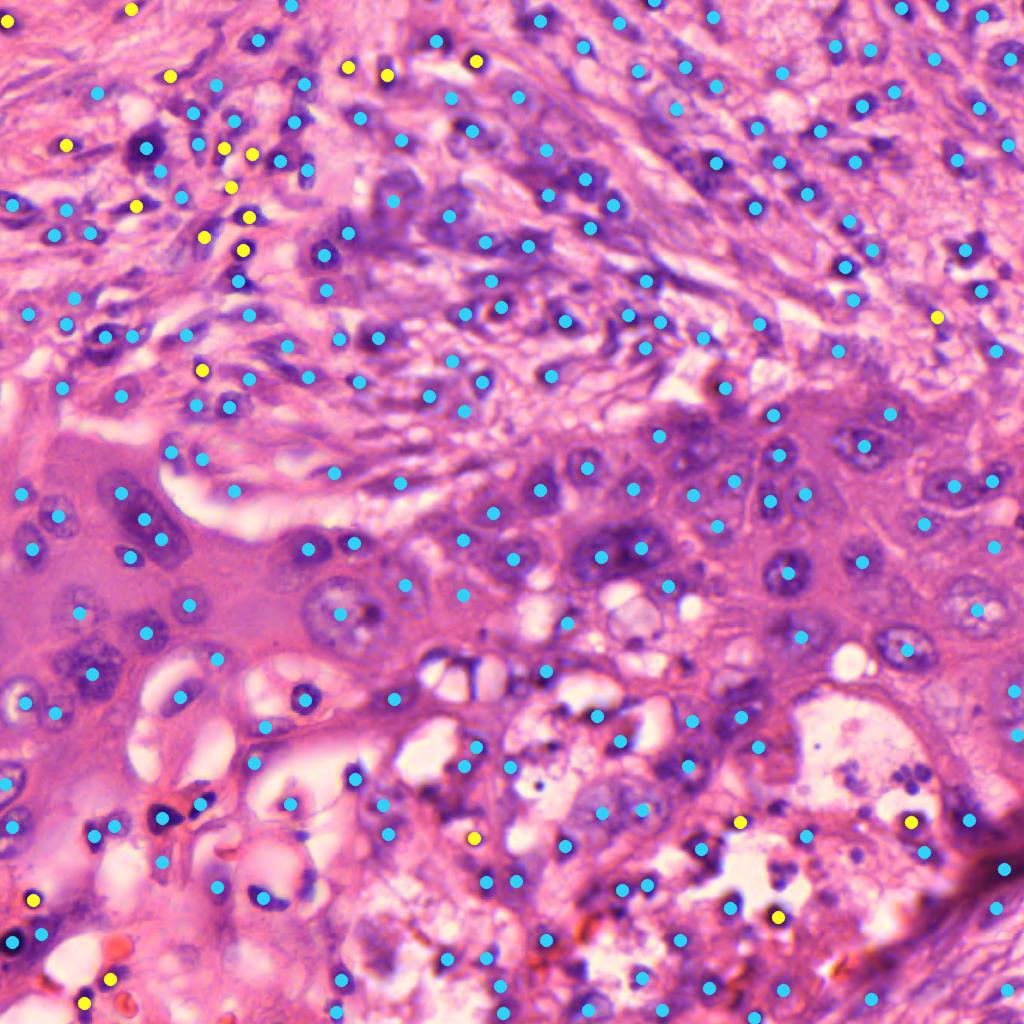}
	\end{minipage}
	\begin{minipage}{0.16\textwidth}
		\includegraphics[width=\linewidth]{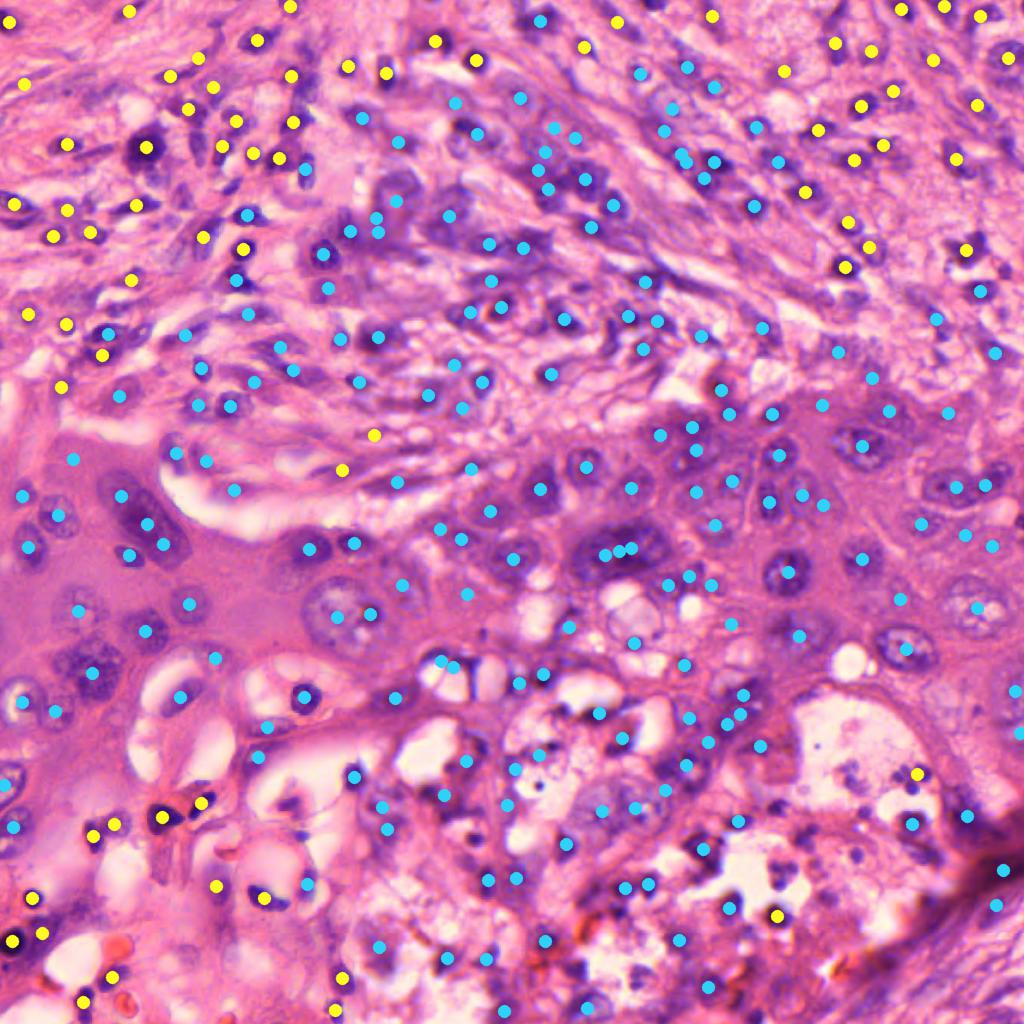}
	\end{minipage}
	\begin{minipage}{0.16\textwidth}
		\includegraphics[width=\linewidth]{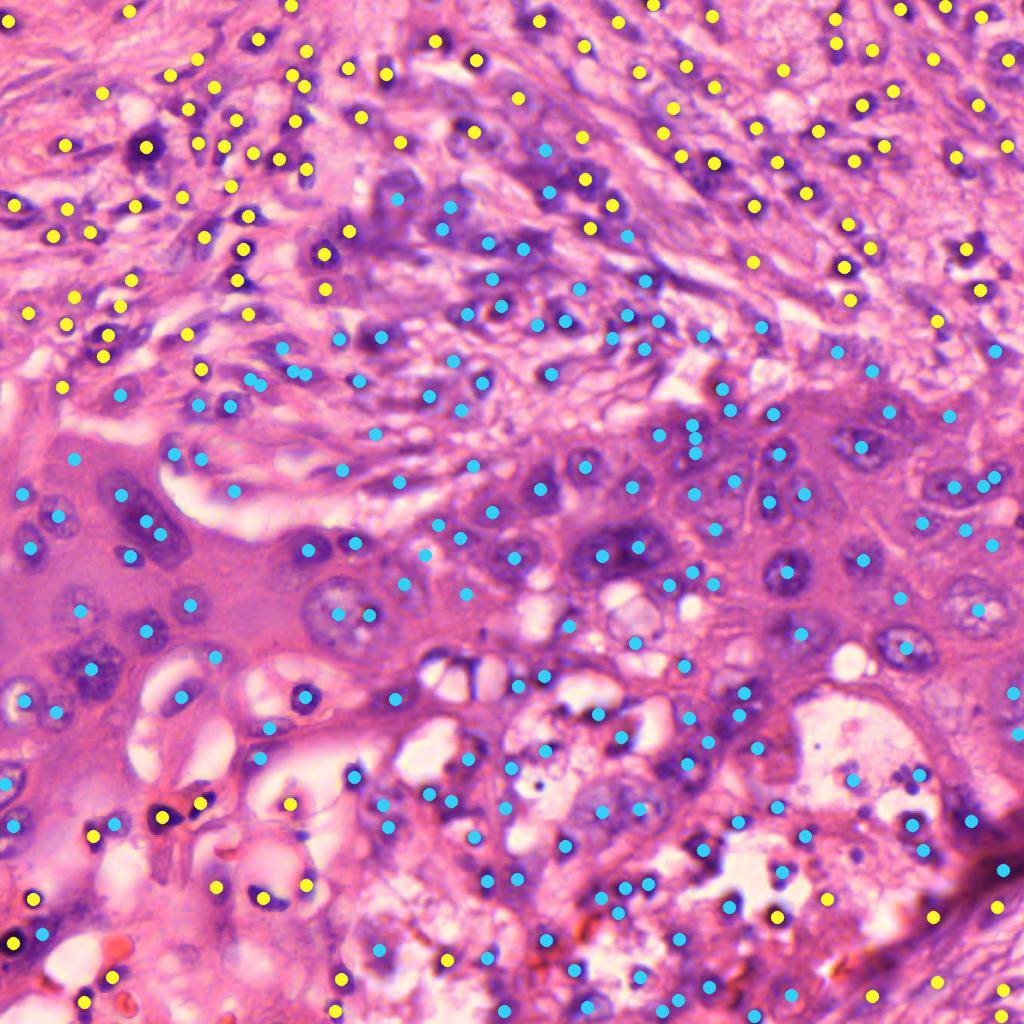}
	\end{minipage}
	\begin{minipage}{0.16\textwidth}
		\includegraphics[width=\linewidth]{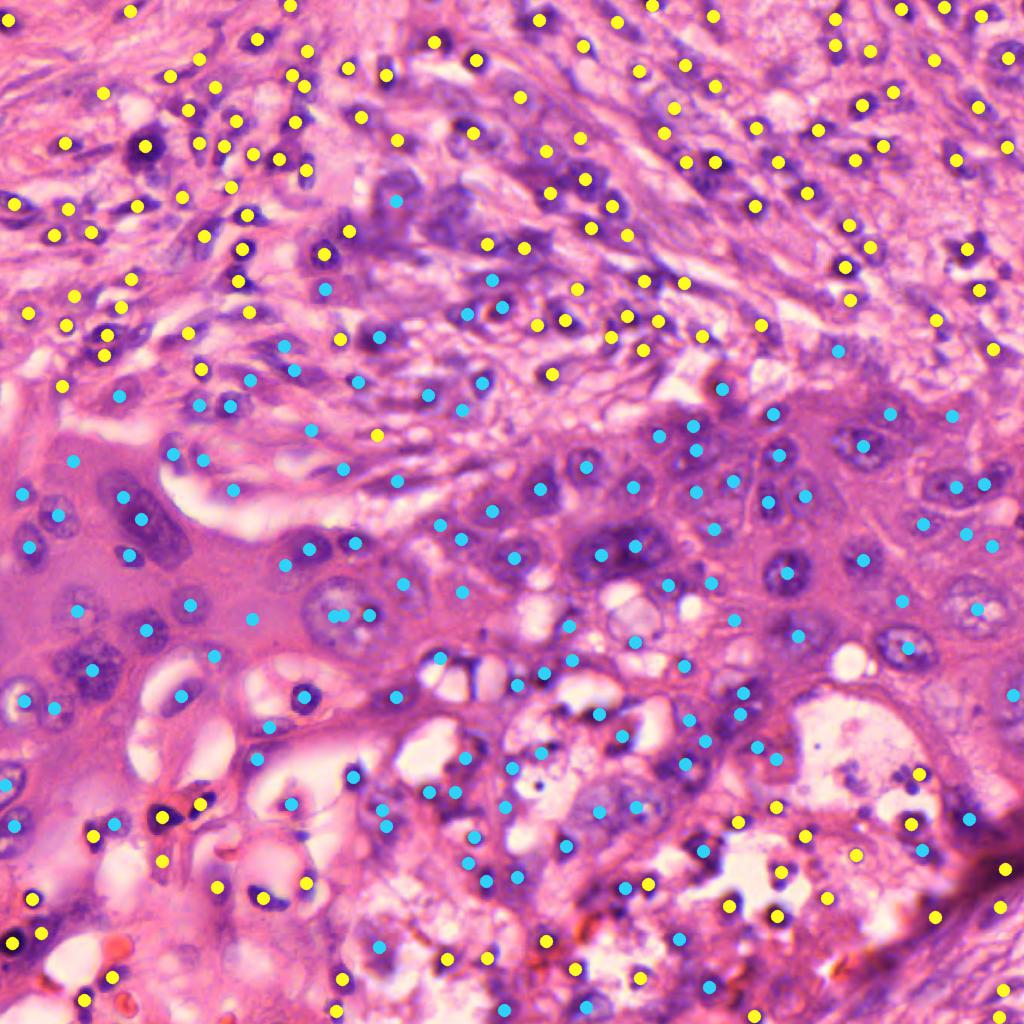}
	\end{minipage} \\
	\begin{minipage}{0.16\textwidth}
		\centering
		\vspace{3pt}
		GT
	\end{minipage}
	\begin{minipage}{0.16\textwidth}
		\centering
		\vspace{3pt}
		$K=1$
	\end{minipage}
	\begin{minipage}{0.16\textwidth}
		\centering
		$K=2^\dag$
	\end{minipage}
	\begin{minipage}{0.16\textwidth}
		\centering
		\vspace{3pt}
		$K=2$
	\end{minipage}
	\begin{minipage}{0.16\textwidth}
		\centering
		\vspace{3pt}
		$K=3$
	\end{minipage}
	\begin{minipage}{0.16\textwidth}
		\centering
		\vspace{3pt}
		$K=4$
	\end{minipage} \\
	\vspace{-5pt}
	\caption{Qualitative comparison results on the OCELOT dataset. The symbol $\dag$ indicates that the predictions come from MFoVCE-Net+, while the rest are from our proposed mFoV DPA-P2PNet. \textcolor{cyan}{$\mdlgblkcircle$} Tumor Cells, \textcolor{yellow}{$\mdlgblkcircle$} Background Cells.}
	\label{fig:vis_mfov}
	\vspace{-5pt}
\end{figure*}

\begin{table*}[t!]
	\centering
	\resizebox{0.8\linewidth}{!}{	
		\begin{tabular}{c|c c c c c|c c c c}
			\toprule[1.5pt]
			\multirow{2}{*}{Metrics} &
			\multicolumn{5}{c|}{\cellcolor{gray!40}\emph{ResNet-50}} & \multicolumn{4}{c}{\cellcolor{gray!40}\emph{ViT-B/16}} \\ \cline{3-10}			 
			& \cellcolor{gray!40}Random & \cellcolor{gray!40}IN & \cellcolor{gray!40}MoCo v2 & \cellcolor{gray!40}SwAV & \cellcolor{gray!40}DINO & \cellcolor{gray!40}Random & \cellcolor{gray!40}IN & \cellcolor{gray!40}DINO & \cellcolor{gray!40}MAE \\
			\bottomrule
			\toprule
			F1 & 52.3 & 55.9 & \textbf{57.4} & 56.4 & \underline{56.6} & 56.7 & 58.0 & \underline{60.8} & \textbf{61.7} \\
			AP & 39.3 & 43.7 & \textbf{45.4} & 44.7 & \underline{45.0} & 44.0 & 47.0 & \underline{49.9} & \textbf{51.6} \\
			\bottomrule[1.5pt]
		\end{tabular}
	}
	\vspace{-5pt}
	\caption{\label{tab:ssl} Downstream evaluation of various SSL algorithms on the large-scale and real-world PD-L1 dataset. IN stands for using ImageNet-supervised pre-trained weights as model initialization.}
	\vspace{-15pt}
\end{table*}

\subsubsection{Effect of mFoV DPA-P2PNet}
We compare the performance of our proposed mFoV DPA-P2PNet with MFoVCE-Net \cite{bai2020multi} and its upgraded version MFoVCE-Net+ \cite{bai2022context} on the OCELOT dataset. These two competitors are designed to improve the cell detection accuracy of DM-based PCD methods by incorporating mFoV patches as model input. The comparison results are shown in Tab.~\ref{tab:mfov}. It can be seen that mFoV DPA-P2PNet that uses 2D transposed convolution for upsampling achieves the highest performance, surpassing the currently SOTA method MFoVCE-Net+ by a remarkable margin of 2.8\% on F1 and 4.9\% on AP. In addition to the inherent shortcomings of DM-based PCD methods, we have identified another factor that results in the inferior performance of these two counterparts. Specifically, they only aggregate the contextual content extracted from a large FoV patch at a single scale, failing to construct sufficiently discriminative features for decoding. By contrast, we injects the plentiful contextual information at multiple scales, as illustrated in Fig.~\ref{fig:mfov}, leading to a significant enhancement of decoding features.

In Tab.~\ref{tab:num_fov}, we further investigate the scalability of mFoV DPA-P2PNet by utilizing more patches with larger FoVs as model input, which has not been previously investigated. Overall, the model performance keeps growing as the number of FoVs increases. As a trade-off, the inference speed gradually slows down. Specifically, when $K$ equals to 4, our approach achieves a substantial performance gain of 5.3\% on F1 and 7.2\% on AP compared to the baseline model that only sees the annotated patch with the smallest FoV. However, the inference efficiency reduces by 68\% as more images need to be processed. To facilitate a straightforward comparison, we visualize the cell detection results under different input conditions in Fig.~\ref{fig:vis_mfov}.

\subsubsection{Ablation Studies} We conduct ablation experiments on the CoNSeP dataset to validate the effectiveness of our proposed modules: multi-scale decoding (MSD) and deformable point proposals (DPP). The results are summarized in Tab.~\ref{tab:msd_dpp}. When using only the MSD, our model achieves a performance gain of 0.7\% on F1 and 1.2\% on AP compared to the original P2PNet. When combined with the DPP, further improvements can be obtained with 1.1\% and 2.4\% on F1 and AP, respectively. In the supplementary material, we empirically confirm that DPP is more favorable to our model than the popular iterative refinement strategy.

\subsubsection{Effect of SSL} To evaluate the transferability of learned weights by SSL, we apply the full fine-tuning protocol to train DPA-P2PNet with a pre-trained backbone on the PD-L1 dataset. Two architectures of backbones including ResNet-50 and ViT-B/16 \cite{dosovitskiy2020image} are tested in our experiments. Moreover, to make the plain ViT model suitable for the dense prediction tasks (e.g., PCD), we introduce ViT-Adapter to inject the image-related inductive biases into the model and construct hierarchical features.

The downstream performance is shown in Table~\ref{tab:ssl}. We observe that supervised ImageNet pre-training is better than training from scratch but lags behind domain-specific SSL pre-training for both ResNet-50 and ViT-B/16 models, which aligns with the conclusions drawn in \cite{kang2023benchmarking}. Of the ResNet-50 based SSL methods, MoCo v2 achieves the most favorable results, outshining the ImageNet-supervised pre-training by 1.5\% on F1 and 1.7\% on AP. Regarding the ViT-B/16 based SSL methods, MAE demonstrates the best performance and it surpasses the ImageNet-supervised pre-training remarkably by 3.7\% on F1 and 4.6\% on AP. We attribute the superiority of MAE over DINO to the fact that contrastive learning primarily captures global relationships, while masked image modeling captures local relationships that is specially beneficial for dense prediction tasks \cite{park2023self}.

\section{Conclusion}
In this study, we present DPA-P2PNet for point-based cell detection. The key improvements of DPA-P2PNet over the prototype model are multi-scale decoding and deformable point proposals, which are designed to promote the utilization of multi-scale information within histopathology images and mitigate the distribution bias between pre-defined point proposals and potential cells, respectively. The ablation studies validate their efficacy and extensive comparison experiments on three histopathology datasets with various staining styles demonstrate the effectiveness and generalization of our proposed DPA-P2PNet model. Based on this, we also design mFoV DPA-P2PNet, the power and scalability of which are validated on the OCELOT dataset. Moreover, we execute the first self-supervised pre-training on large scale IHC image data and evaluate the efficacy of various SSL methods on the PCD task specially. We hope that our work can advance the development of computational pathology community.



\bibliography{aaai24}

\end{document}


\maketitle

\section{Datasets}
We present the detailed cell categories of each dataset as follows:
\begin{itemize}
	\item \textbf{CoNSeP} \cite{abousamra2021multi}: epithelial, stromal cells and inflammatory cells.
	\item \textbf{BCData} \cite{huang2020bcdata}: positive/negative tumor cells.
	\item \textbf{PD-L1}: positive/negative tumor cells, positive/negative histocytes, positive/negative lymphocytesfibrous interstitial cells, other inflammatory cells, normal alveolar cells and others.
	\item \textbf{OCELOT} \cite{ryu2023ocelot}: background cells and tumor cells.
\end{itemize}

\section{Experiments}
We compare the suitability of deformable point proposals (DPP) with the popular iterative refinement (IR) strategy for our model on the CoNSeP dataset. The experimental results are exhibited in Tab.~\ref{tab:ir}.
It can be seen that DPP contributes to performance improvement, whereas IR exhibits an increasingly detrimental effect as the number of stages increases. We attribute the performance degradation caused by IR to the severe foreground-background imbalance.
\begin{table}[h]
	\centering
	\resizebox{0.7\linewidth}{!}{	
		\begin{tabular}{c |c c }
			\toprule[1.5pt]
			& F1 & AP \\
			\bottomrule
			\toprule
			Baseline & 70.7 & 61.7 \\
			DPP & 71.1 (\textcolor{green}{+0.4}) & 62.9 (\textcolor{green}{+1.2}) \\
			2-stage IR & 69.4 (\textcolor{red}{-1.3}) & 59.1 (\textcolor{red}{-2.6}) \\
			3-stage IR & 69.3 (\textcolor{red}{-1.4}) & 58.7 (\textcolor{red}{-3.0}) \\
			\bottomrule[1.5pt]
		\end{tabular}
	}
	\caption{\label{tab:ir} Comparison with the iterative refinement strategy. The baseline outcomes are obtained by incorporating only the multi-scale decoding strategy into the original P2PNet model.}
\end{table}

\bibliography{aaai24}